# BUILDING A TEMPERATURE FORECASTING MODEL FOR THE CITY WITH THE REGRESSION NEURAL NETWORK (RNN)

Tran Nguyen Phuc[1], Duong Thi Thuy Nga[2], Tran Duy Thanh[3]

[1]*Software engineer, Egopulse JSC, Viet Nam*
[2]*Ph.D Duong Thi Thuy Nga, Faculty of environment, Ho Chi Minh University of Science, Viet Nam*
[3]*MSc Tran Duy Thanh, Faculty of information system, University of Economics and Law, Viet Nam*
Nguyenphuc9119@gmail.com[1], ngadtt@hcmus.edu.vn[*2], thanhtd@uel.edu.vn[3]

**ABSTRACT** In recent years, according to some studies of environmental organizations in the world and Viet Nam show that the weather change is quite complex. Global warming has become a serious problem in modern world which is the concern of scientists. Last century, it's difdicult for us to forcast the weather. Why can't scientists accurately predict the weather? The reason is that the weather monitoring stations were missing and technological limitations. Therefore, it's hard to collect data for the construction of predictive models to make accurate simulations. In Viet Nam, the research for weather forecast models have been applied recently. They have been developed since 2000. Along with the development of computer science, mathematical models are built and applied to machine learning techniques that help to build accurate and more reliable predictive models. The results of research and solutions for the application of urban temperature forecasting with the recurrent neural network is going to be summarized in this article.

## 1. Introduction

The weather change is making global warming. The temperature affects the production and daily life. It is necessary to apply the results from science to build the temperature forecasting applications. It helps to take the initiative in monitoring environmental temperature. Besides, it provides scientific reference data for experts to investigate and analyze environmental temperature, from that points, experts can give some methods to deal with the problems.

In the past decade, many researchers in the world have been trying to solve the forecast of weather by using a statistical model, including machine learning techniques that has been reported to reach many successful results.

From the above assumptions, I believe that building and developing a weather forecasting model has both been academic and realistic. Although the approach needs to be invested in time as well as data collection for the system to be relatively large, it is a long term to allow us to proactively expand the system with other parameters such as humidity, wind, vision, … for research and practical applications.

It is undeniable that nowadays people have studied and widely applied information technology and machine learning techniques in various fields such as: gold price forecast, stock market prediction, planning for economic development… In Viet Nam, although the research and appilication of machine learning have been developed recently, they are gradually forming and improving with significant achievements.

The benefits of applying machine learning techniques:

- More and more information come from different sources, so people must process and filter the information that becomes difficult and costly. The appication of machine learning can help us classify, analyze, and extract information from data sources quickly, save time and reduce our efforts.

- Without experts, machine learning can help create analytics applications and make decisions from data collected.

- The algorithm of machine learning can help analyze and populate incomplete, inaccurate data.

- Machine learning techniques can help predict missing data and validate data.

- Machine learning can help build a training automation system (AI) and mining relationships between data collected.

The advantage of machine learning is that it generates rules, which can be modified, or trained in a certain limit. The information is presented follow a structure including four-level which called the pyramid knowledge. The lowest level is a large data which is saved information, it is very difficult for us to understand. So, it is necessary to extract useful information from the data warehouse by using preprocessing methods. The results of the processing will create the second level. The third level of knowledge is often created by rules generated.

The processing of machine learning consists of two steps:

- **The learning period**: when the system analyzes data, they recognize the relationships between data (linear and nonlinear). The results can be as: groups of objects in the classes, generated rules, predicting classes for new objects.
- **Testing period**: The relationships (rules, classes,…) generated must be re-tested by validation function on a part of the training data or on a large data set.

In general, the machine uses a limit set of data called training data. This data set contains data in a format that machine can understand. However, the training data set always have a limitation so we can't make sure all the data was learned correctly.

The regression network (RNN) is an outstanding algorithm that has been used in building predictive models from data collected in the past. It is shown by the results of scientific works in the past with high accuracy.

## 2. Related Works

**General about regression and approach**

Regression functions or mean (conditional expectation) is one of the most important theory, which applied in many fields: Econometric, solving the problem of analytic and prediction, ... This section will present some approaches.

*Approach from reality:* each other in sum $\Omega$, we observe two variables (two standards) X, Y. $\Omega x$ symbol is a set of elements of $\Omega$ that having the same value X=x. If limit observe Y on set $\Omega x$, we have $Y_X$, it means $Y_X = Y \mid \Omega x$ (limit of Y on set $\Omega x$). Then, the average value of $Y_X$ is $EY_X$ that called mean (conditional expectation) of Y with condition X=x and signed E (Y|X = x). So, we have:

$$E(Y|X = x) = EY_X$$

When x changes, we have f(x)= E (Y|X = x). Sign E(Y|X) = f(X), is random variable, select the value E (Y|X = x) when X= x, and it called conditional expectation or average condition of Y by X. It's also called that overall regression function of Y by X. singed PRF of Y by X.

*Approach from the mathematical model:* the example we have X, Y are the random variables on a space of base probability ($\Omega$, F, P). In there, Y have moment at level 1 with limited. Set B is $\sigma$ – algebra Borel on set R; $B(X) = X^{-1}(B)$ is $\sigma$ - algebra generated by X; $P_{B(X)} = P \mid B(X)$. Then $\varphi(A) = \int_A Y dP$, $A \in B(X)$ is an extrapolation measure on B(X), absolute continuity by measure $P_{B(X)}$. Radon's derivative of $\varphi$ by $P_{B(X)}$: $\frac{d\varphi}{dP_{B(x)}}$ is the function B(X)- measurable and called is a conditional expectation of Y according to X, then we have:

$$E(Y|X) = \frac{d\varphi}{dP_{B(x)}}$$

*Approach from general requirements:* when observing X, Y in statistical dependence, depend on the information of X, we want to find a random variable (depend on X or a function of X) that has the best approximate value for Y by minima of mean squared deviation. The random variable that we need to find is projection (perpendicular) of Y on space of functions depend on X, or called is regression function of Y by X, signed E(Y|X).

*Example*: Overall are all family in an area. Each family we consider two standards: X is monthly incoming; Y is monthly spending. Then $\Omega_X$ is a set of all family have the same income X = x; E (Y|X=x) = $Ey_X$ mean average spend of family in area have the same income X = x.

**Machine learning algorithms**

Machine learning algorithms are divided into three types: supervisor learning, unsupervised learning and half supervision learning.

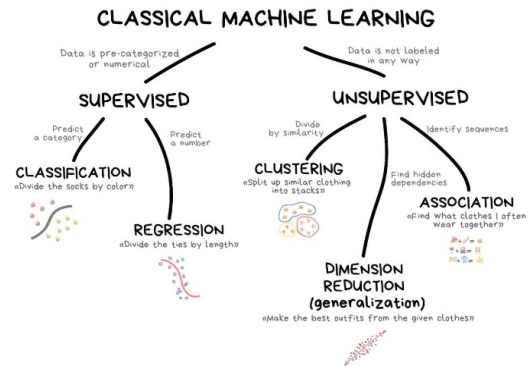

*A branch of machine learning – Source: Internet*

*Supervisor learning*

This type is learning from the data set that are provided where machine learning techniques can build a way to define data classes. The system must find a description for each class (feature layer of data). Thus, people can use rules that are formed during learning and classifying to be able to use classes forecast later.

The supervisor learning algorithm has a set of data train include M pairs:

$$S = \{(x_i, c_j) \, i=1, …, M; j=1, …, C\}$$

These training pairs are called samples, in which:

- $x_i$ is vector n dimension, also known as feature vector
- $c_j$ is the first class known.

The supervisor learning algorithm will try to find all features of the hypothesis, called H. For one or more theories, the best estimate is not known exactly $f: x \rightarrow c$. For classify, we can use hypothesis such as a class criterion. Machine learning algorithms find hypotheses by discovering common features of example that show each class. The results are often in the form of rule (if... then).

Depending on the level of the machine learning algorithm, they have the following surveillance models:

- *Learning like rote:* the system is always "taught" by all the right rules, and then learning with convergence.
- *Learning by analogy:* the system is taught feedback for a similar task, but it's undefined. So, the system must calibrate feedback by creating a new law that can apply to new cases.
- *Learning by based on the case:* in this case the learning system stores all cases, along with their output results. When it finds a new case, it' s going to try to edit the new case that previously stored.
- *Learning based on interpretation:* the system will analyze the set of solutions to indicate why each method is successful or unsuccessful. After these explanations are created, they will be used to solve new problems.

*Unsupervised learning*

This is learning from observation and discovery. The mining data system is applied to objects but there is no class defined before, but it must be self - aware of patterns and recognize patterns. The system generated a set of classes, each with a set of samples explored in the data set.

In cases where there are few, or almost no knowledge of input data, then an unsupervised learning system explores the classes of data, by finding the attributes, the common characteristics of forms forming data sets.

A supervisor learning algorithm can always transform into an unsupervised learning algorithm. For a problem that data patterns are described by n characteristics, they can run the algorithm for monitoring n - times, each time with a different characteristic, which is a class attribute, which we are predicting. The result is that n class criteria, with the hope that at least one of the n class is true.

*Semi supervised learning*

Semi supervised learning is algorithms integrated from supervised learning and unsupervised learning. Semi supervised learning uses the advantages of supervised learning and unsupervised learning then eliminating common defects on these two types.

The first time, computers were used in meteorology in 1950. Arithmetic calculator and computer (ENIAC) were utilized to imitate the weather condition in Aberdeen, Maryland, Unites States. Then, the strong development of computer science has produced a lot of predictive algorithms and is being developed as strongly as: support vector machine (SVM), k - nearest neighbors (KNN), Recursive neural network (RNN), Convolutional neural network (CNN), decision tree, random forest, … to help build more accurate and reliable forecasting models.

*General about Recursive neural network*

The regression network (RNN) is an algorithm that has a lot of attention in building predictive models, from data collected in the past. Because of forecast results in the previous scientific works that show good and high accuracy. The main idea of RNN is to use the sequence of information and give the following information. The input and output data of traditional networks (as CNN, ANN, …) is almost entirely independent of each other, which means they have no connection to each other. In the other hand, RNN is very suitable for input type and linked output as a series, so the selection of RNN models suggests that the availability of an application is in the temperature forecast. RNN is called the regression network because the algorithm performs the same task for all elements of a series with the output of the dependent algorithm depending on the previous calculations. From the above remarks, within the scope of this article, the neural network algorithm has been selected for research and presentation of building the temperature forecasting model.

The neural network model combines the fuzzy theory and the neural network model that takes advantage of advantages such as the ability to approximate a continuous function with the accuracy, exploit the ability to handle such knowledge as human (fuzzy theory). This model is appropriate because of the requirement that the input and output are dependent on time. The fuzzy neural network appears to be highly effective for applications such as a time series forecast, identification, and control of nonlinear systems.

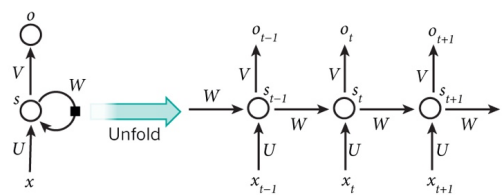

*Neural network model RNN, source: internet*

The illustration model depicts the content implementations of the RNN. Calculations at step n are all dependent on the previous steps that (n - 1 return to previous) as base data. The first layer will be received from the input data and the next layers (hidden layer) will receive in the previous layer. The neural network also supports the implementation of function (activation function) to determine when to continue or stop regression.

The neural network model consists of 3 classes:

- *The first layer:* class of input has N data input (input layer)
- *The second layer:* Called are hidden layers that contain member functions. The nodes in this class perform fuzzy, which is used to compute the values of member function values in a distribution function.
- *The third layer:* layer of data output, input data from calculations through hidden layers and extract forecasts (output layer).

Example we have the data x(t) and y(t), that is the input and output of the corresponding time series, we have a neural network that can divide into three groups of $W_{ih}$, $W_{hh}$ and $W_{oh}$. The following model represents the simulation of the operation of the regression neural network:

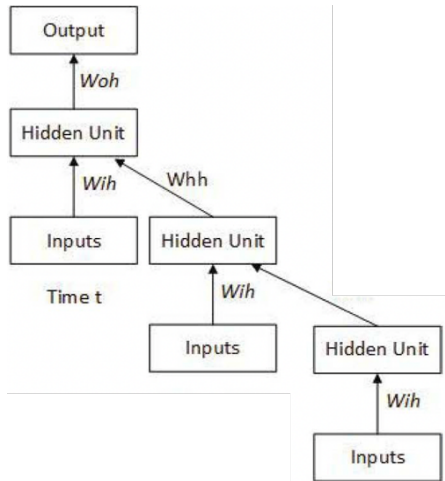

*The recursive neural network activity model*

With:

- $W_{ih}$: input data acquisition layer as weights
- $W_{hh}$: the hidden layers of the neural network
- $W_{ho}$: the layer to extract output data that corresponds to the input data

## 3. Proposed Method

**Objectives**

- Building a temperature forecasting model for the city with regression neural network (RNN).
- Developing the application for forecasting temperature using the model that built from the result.

**Methods**

- Theoretical and empirical research methods: research about machine learning techniques and its applications in the real-life. Data analysis and select algorithms that match with the data, thereby building the temperature forecasting model for the city and apply suitable technical to the collection data set.
- Experimental method: analyze and divide data into two parts. A part will use for training data with RNN, after that use the trained model for validation from the other data. After that compare the forecasting results with practical results in the testing data set.

**Data collection**

In the article using data provided by the website meteoblue.com as the source for the base data to build the temperature forecasting model, in the article will build the forecasting model in Ho Chi Minh city, Vietnam.

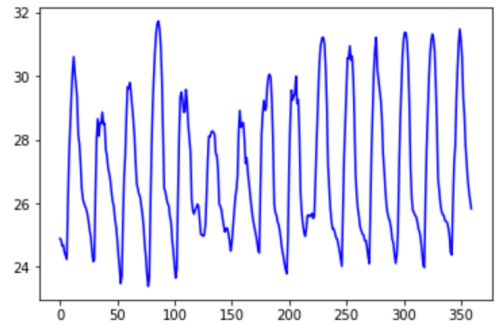

*A part of the collected data*

The data provided includes data fields such as year, month, day, hour, minute, minute, temperature, total precipitation, wind speed, wind direction. However, we're just interested in the **temperature** field to use data sources for training.

*The visual of simulate temperature in a series* continuous *time (360 hours)*

The visual shows temperature on a continuous period, we notice that the temperature change tends to be cyclic, the largest temperature margins usually fall into the time period [11 AM - 2 PM] and the lowest usually in time [12 AM – 3 AM] of the same day.

The data set will be divided into 2 parts (70% and 30%) and is named corresponding to the data set for (training data), including 70%, the other 30% as the data for testing after obtaining the results from the machine learning machine.

The data set will be created in the following guidelines:

- Data is divided into two columns X and Y
- Data at line n + 1: X will receive the value of Y at line n
- Data of 1 day corresponds to 24 pairs of values X and Y (corresponds to each hour)

- Data is grouped into blocks that each block corresponds to 24 hours

| X | Y |
|---|---|
| 25.63 | 25.38 |
| 25.38 | 25.20 |
| 25.20 | 25.01 |
| 25.01 | 24.81 |
| 24.81 | 24.51 |

*The table show sample data created by the rule that described above*

## 4. Experimental Results

*The architect of system and the results of testing*

The program is installed and tested based on the neural network model RNN, using the result generated from machine learning techniques with the neural RNN for the temperature forecasting model as bellow.

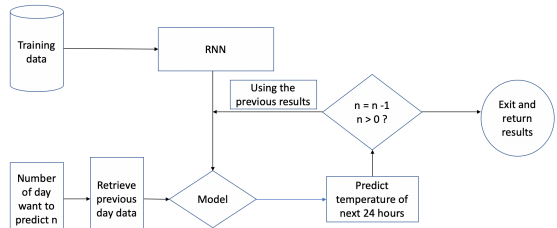

*Diagram of handling in the system*

Testing results from the temperature forecasting model with RNN:

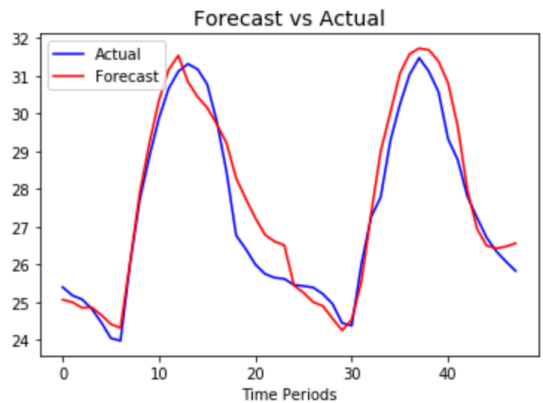

*The result of predicting (red) compare with the actual value (blue) in the next 48 hours*

The testing result of the temperature forecasting with RNN showed that the predicted results are intuitive for the expected data for the next 48 hours, with the exact 96.2% ratio measured by the measure:

**Mean absolute error (MAE)** = $\frac{\sum_{i=0}^{n}\left|\frac{Pi-Ti}{Ti}\right|}{n}$

With:
- n is the total amount of forecast temperature
- Pi is the forecast temperature at position i
- Ti is the temperature of the position i in reality

Experimental environment and parameters

Experimental environments are run on Google Colab with libraries: *Tensorflow, Pandas, Numpy*.

| Parameter | Value |
|---|---|
| rnn_size | 100 |
| learning_rate | 0.001 |
| dropout_keep_prob | 0.5 |
| Epochs | 1000 |
| rnn_cells | num_unit=100, activation=tf.nn.relu |

*The table describes parameters to build and test model*

## 5. Conclusion

The temperature forecasting model shows the positive and acceptable results with the application of recurrent neural network to the building of predictive models. The basic solution has integrated and responsive to the future temperature forecast based on historical data. However, to be able to examine and accurately assess the scope of the model, we must test and use of the long-term.

The most important and main contribution of this research is to propose models of machine learning, which based on using the neural network for forecasting the temperature in Viet Nam. moreover, it's a premise to develop the weather forecasting model in the next research. It is, thus, contributing to create an information for temperature forecasting channel, which helps scientists, experts have more reference data in assessing, giving preventive measures to climate change.

In addition, it appropriate for the application of forecasting temperature in the near future (2 - 7 days) that helps users have more information about the temperature, which extends and proposes the development of the national data on the temperature of the province/city in Viet Nam, which facilitates the research of temperature change in Viet Nam.

Soon, the model will continue to research and expand with the parameters affecting temperatures such as: urban planning density, rainfall, humidity, wind speed, …